# French Word Recognition through a Quick Survey on Recurrent Neural Networks Using Long-Short Term Memory RNN-LSTM

Saman Sarraf*

*Department of Electrical and Computer Engineering, McMaster University, Hamilton, Ontario, Canada, The Institute of Electrical and Electronics Engineers, Senior Member IEEE*

*Email: samansarraf@ieee.org*

**Abstract**

Optical character recognition (OCR) is a fundamental problem in computer vision. Research studies have shown significant progress in classifying printed characters using deep learning-based methods and topologies. Among current algorithms, recurrent neural networks with long-short term memory blocks called RNN-LSTM have provided the highest performance in terms of accuracy rate. Using the top 5,000 French words collected from the internet including all signs and accents, RNN-LSTM models were trained and tested for several cases. Six fonts were used to generate OCR samples and an additional dataset that included all samples from these six fonts was prepared for training and testing purposes. The trained RNN-LSTM models were tested and achieved the accuracy rates of 99.98798% and 99.91889% for edit distance and sequence error, respectively. An accurate preprocessing followed by height normalization (standardization methods in deep learning) enabled the RNN-LSTM model to be trained in the most efficient way. This machine learning work also revealed the robustness of RNN-LSTM topology to recognize printed characters.

## 1. Introduction

### *1.1. Recurrent Neural Network (RNN)*

Recurrent neural networks are effective architectures for sequence learning tasks where the data is highly correlated along a single axis. This axis usually corresponds to time domain, or in some cases one-dimensional space; for example, a given sequence of proteins.

------------------------------------------------------------------------

* Corresponding author.

Some of the properties that make RNNs suitable for sequence learning - such as robustness to input warping and the ability to learn which context to use - are also desirable in domains with more than one spatiotemporal dimension. However, standard RNNs are inherently one-dimensional, and therefore poorly suited to multidimensional data. Multidimensional RNN expand the application of this architecture to computer vision, video processing and medical imaging, as it enables researchers to use multidimensional data [1, 2]. Recurrent neural networks were originally developed as a way of extending feedforward neural networks to sequential data. The addition of recurrent connections allows RNNs to make use of previous context, and makes them more robust to warping along the time axis than non-recursive models. Access to contextual information and robustness to warping are important when dealing with multidimensional data. The standard RNN architectures are inherently one-dimensional, meaning that in order to use them for multidimensional tasks; the data must be preprocessed to one dimension, e.g. by presenting one vertical line of an image at a time to the network. Perhaps the best-known use of neural networks for multidimensional data has been the application of convolutional networks to image processing tasks such as digit recognition [3, 4, 5, 6]. Various statistical models have been proposed for multidimensional data, notably multidimensional hidden Markov models. However, multidimensional HMMs suffer from two serious drawbacks: (1) the time required to run the Viterbi algorithm (and thereby calculate the optimal state sequences) grows exponentially with the size of the data exemplars, and (2) the number of transition probabilities, and hence the required memory, grows exponentially with the data dimensionality.

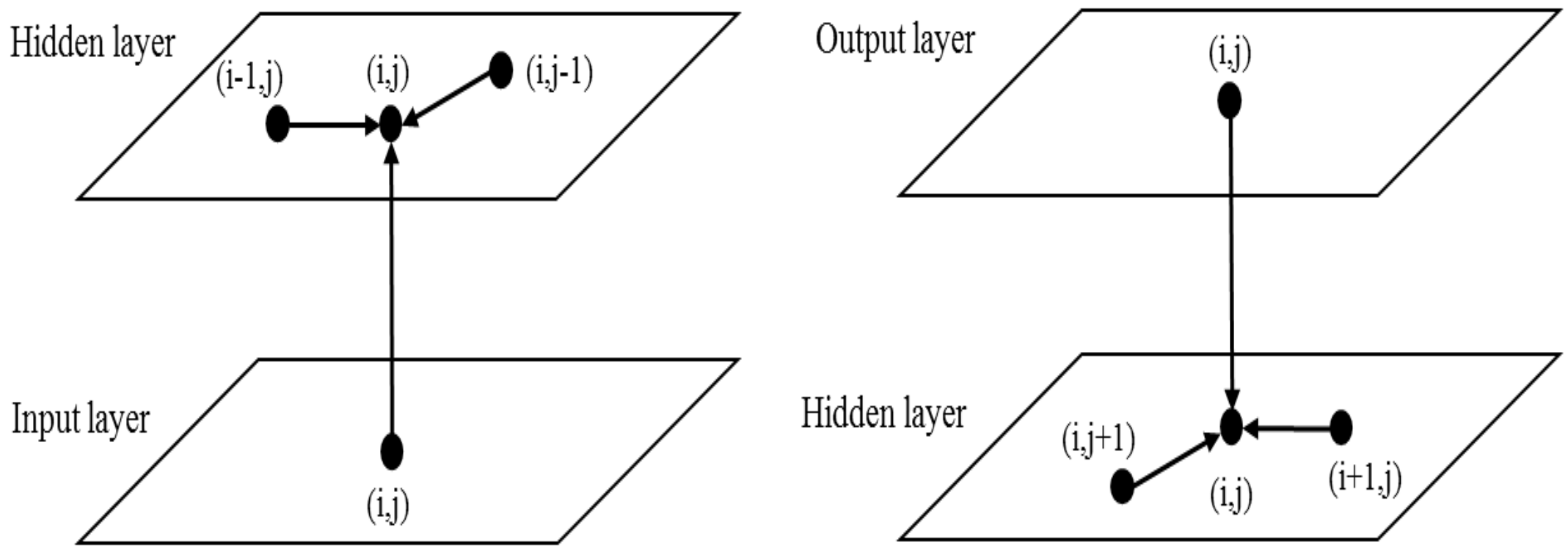

**Figure 1:** Multidimensional Recurrent Neural Networks forward pass (left) and MDRNN backward pass (right)

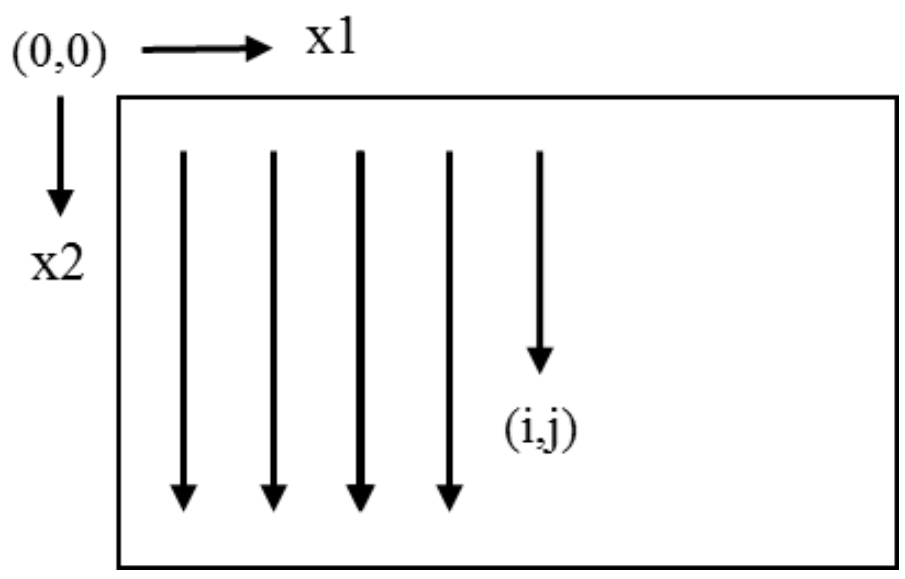

**Figure 2:** Sequence ordering of 2D Data. The MDRNN forward pass starts at the origin and follows the direction of the arrows. The point (i,j) is never reached before both (i-1,j) and (i,j-1).

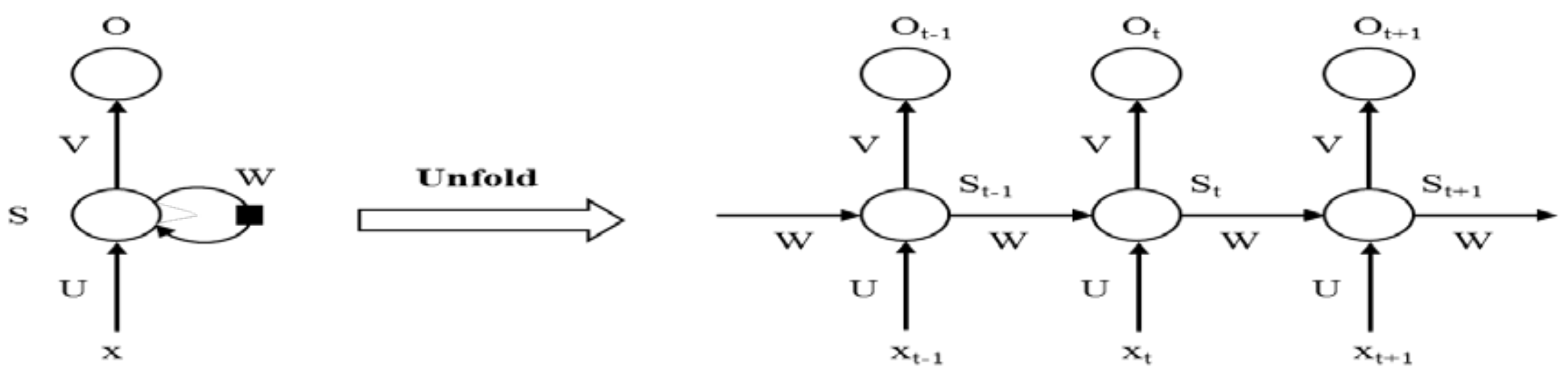

**Figure 3:** A recurrent neural network and the unfolding in time of the computation involved in its forward computation. Source: Nature

### *1.2. Long-Short Term Memory (LSTM)*

The most important characteristic of recurrent neural networks is use of contextual information between input and output sequences. However, accessing the information flow in the RNN architectures is limited in practice, due to exponential decay of the influence of inputs on hidden layers around the recurrent connections. This issue is called the vanishing gradient problem [6]. In past decades, research groups proposed solutions for overcoming vanishing problems [3]. The Long-short Term Memory (LSTM) is formed by a set of recurrently connected components called memory blocks. Each memory block often contains a self-connected memory cell, input, output and forget gates that enable updating of the given block. Figure 3 images a single-memory block of LSTM. The multiplicative gates allow LSTM memory cells to store and access information over long periods of time, thereby mitigating the vanishing gradient problem. As long as the input gate remains closed (i.e. has an activation near 0), the activation of the cell will not be overwritten by the new inputs arriving in the network and can therefore be made available to the net much later in the sequence by opening the output gate.

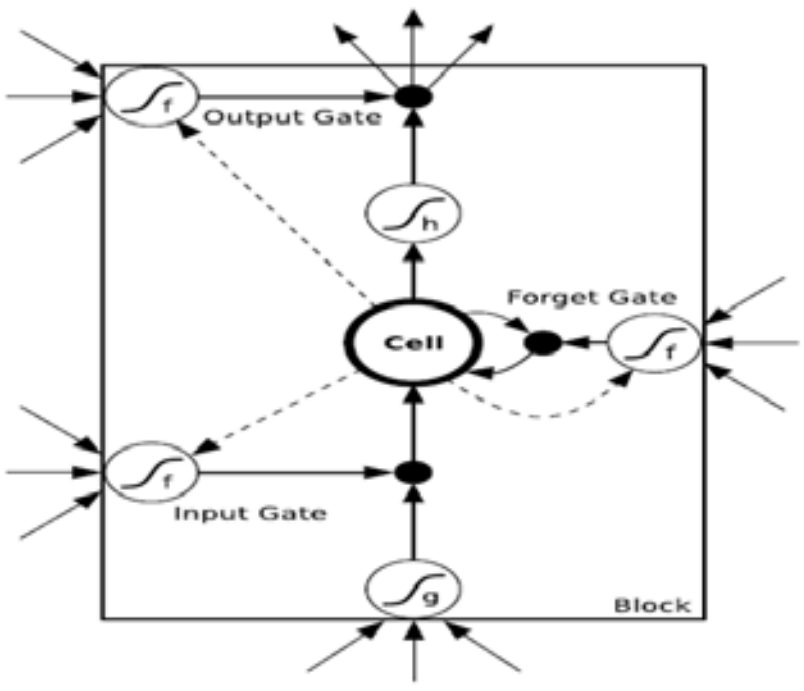

**Figure 4:** LSTM memory block with one cell is shown. The three gates are nonlinear summation units that collect activations from inside and outside the block and control the activation of the cell via multiplications.

The input and output gate multiply the input and output of the cell while the forget gate multiplies the cell's previous state. No activation function is applied within a given cell. The gate activation function is usually the logistic sigmoid, therefore the gate activations are between zero and one. The cell input and output activation functions are TANH or logistic sigmoid. The weighted peephole connections from the cell to the gate are shown with dashed lines. All other connections within the block are unweighted. The only outputs from the block to the rest of the network emanate from the output gate multiplication.

Using LSTM as the network architecture in a bidirectional recurrent neural network yields bidirectional LSTM. Bidirectional LSTM provides access to long-range context in both input directions [1, 5]. For most LSTM networks, the computational complexity in terms of time is O (W) for feedforward and feedback operations. This means that the bidirectional networks and the LSTM networks take significantly no more time per training epoch than the unidirectional or RNN or multi-layer perceptron networks.

## 2. Method

### *2.1. Data collection and preparation*

Using online resources, the top 15,000 French words including all French letters were selected. Figure 5 shows the characters' distribution in the dataset. The distribution shows that the focus of data collection was mostly on lowercase letters. Among popular MS fonts, the more frequently used Arial, Calibri, Cambria, Georgia, LucidaFax, and Times New Roman were selected to create the imaging samples. The words were then split into three columns and roughly 30 lines per page, and text files of each font were converted to PDF with DPI 300 followed by JPG conversion. Also, an additional dataset including all the fonts' samples was generated containing 90,000 words.

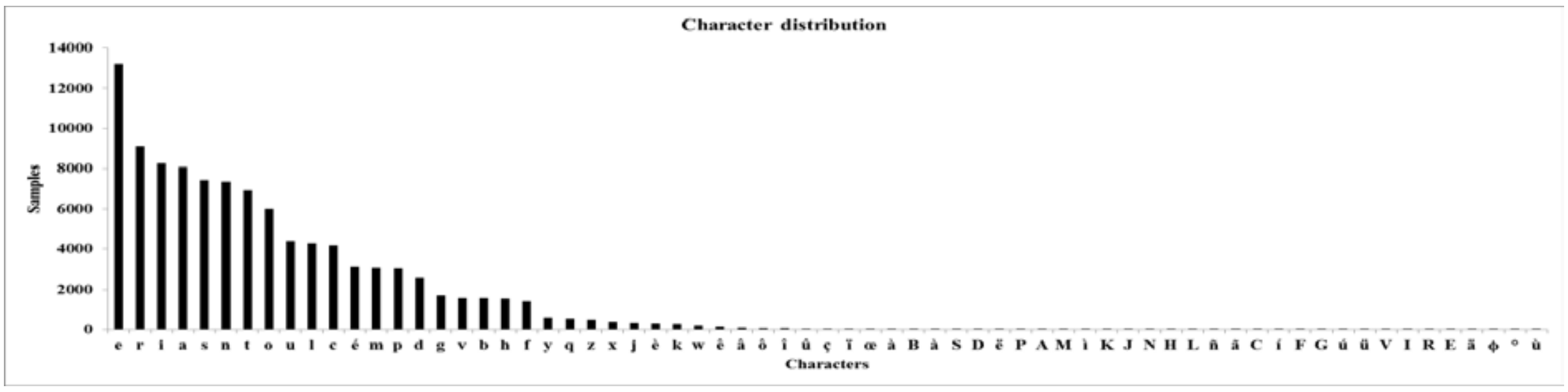

**Figure 5:** Character-level data distribution of the words used to train and test RNN models

### *2.2. Image preprocessing*

Global noise removal was performed using a median filter with a window size of three. The median filter replaced a pixel by the median of the pixels in the window. This removed any potential and isolated noise occurring during data generation such as image conversion or document scanning. In the next step, the native RGB images were converted to the gray-scaled images for further processing. The samples were binarized using the Otsu thresholding algorithm [7]. Otsu's method is an iterative algorithm to perform clustering-based image thresholding. In the final step of preprocessing, the samples were inverted to adapt to next image processing modules.

### *2.3. Image segmentation*

The preprocessed page-level data were passed through the line and word segmentation modules. Firstly, the sum of pixel intensities for each row on a given page was calculated. Given that between each line of words, the sum of intensities must be almost zero, each line's coordinates were estimated. In the next step, the segmented lines were passed through the word segmentation module. The sum of each column in a given segmented line was measured. The same zero finder method as mentioned above was utilized to estimate the coordinates to each word. As the size of each letter is different, the estimated coordinates might generate the segmented words with different zero padding. Therefore, each segmented word was re-segmented to only have one zero-padded pixel from any direction. The enabled the samples to have a standard format that facilitate the learning processing of classifiers.

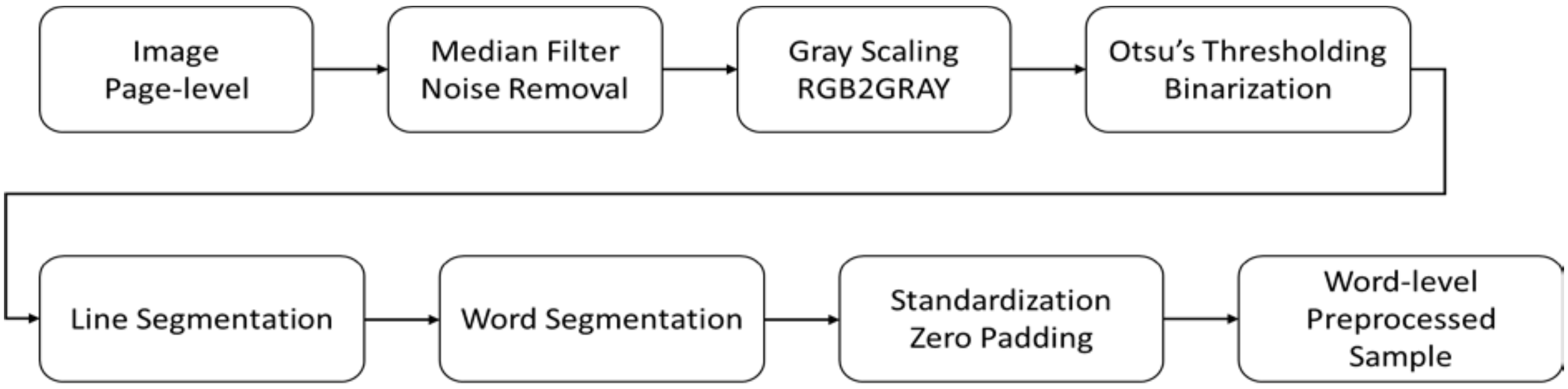

**Figure 6:** Image processing pipeline

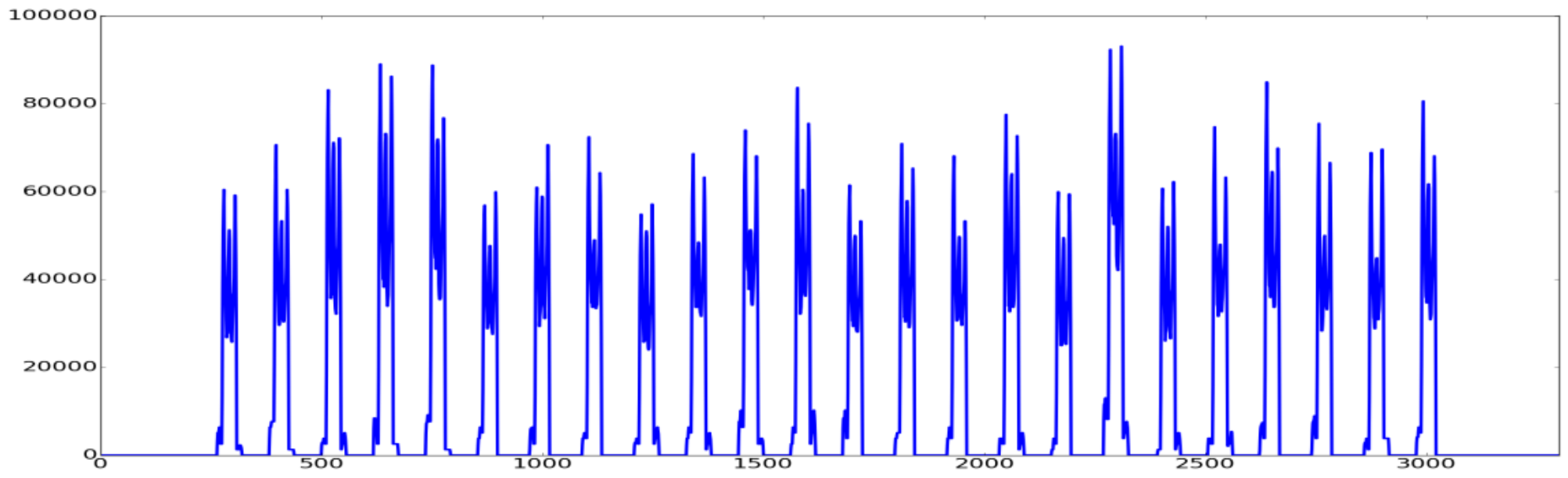

**Figure 7:** Sum of pixel intensities over rows for a given page. Based on the zero values between peaks, the figures shows there are 25 lines in the page.

### *2.4. RNNLIB - Deep learning framework*

Classification of word-level data was performed using an open source machine learning framework called RNNLIB. This framework is a recurrent neural network (RNN) library for sequence labeling tasks such as handwriting or speech recognition. Various types of machine learning architectures are available in this library including LSTM, traditional neural networks, multilayer perceptrons, and standard recurrent networks with nonlinear hidden units. Additionally, RNNLIB provides bidirectional LSTM that enables a given training process to consider long-range contextual information in all input directions. RNNLIB also provides the capability of using connectionist temporal classification (CTC) enabling a given system to transcribe unsegmented sequence data. Furthermore, multidimensional recurrent neural networks that is provided by RNNLIB extends the system to data with more than one spatiotemporal dimension, allowing researchers to use 2D, 3D, or 4D data such as images, videos, and functional MRI [8].

### *2.5. Label Error Rate (LER)*

Performance of classification is measured using various metrics. In printed characters and handwriting recognition, label error rate (also called edit distance error) is the standard metric for evaluating character level classification performance. Given a test set $\mathbf{S}' \subset \mathbf{D}_{\mathbf{x}\times\mathbf{z}}$ that is independent of $\mathbf{S}$, the label error rate (LER) defined by Equation 1:

$$LER(h, S') = \frac{1}{|S'|} \sum_{(x,z)\in S'} \frac{ED(h(x), z)}{|z|} \quad (1)$$

Where h is a temporal classifier as the mean normalized edit distance between its classification; the targets on $\mathbf{S}'$ $\mathbf{ED}(\mathbf{p}, \mathbf{q})$ is the edit distance between two sequences; **p** and **q** which is the minimum number of insertions, substitutions, and deletions require to change **p** into **q.** .This is a natural measure for tasks (such as speech or handwriting recognition) where the aim is to minimize the rate of transcription mistakes.

### *2.6. Connectionist Temporal Classification - CTC Error*

Alex Graves and his colleagues [5] developed neural networks called Connectionist Temporal Classification, where a recurrent neural network is used for CTC. The most important step in this development was to transform the network outputs into a conditional probability distribution over label sequences. The network was then used as a classifier by the selecting the most probable labeling for a given input sequence. In this work and in each epoch of training, the LER and CTC error measured the best model based on each metric reported as the final result of the training.

## 3. Results and Discussion

The machine learning-based solution in this work was designed by randomly dividing each of seven datasets mentioned earlier into training and testing datasets including 80% and 20% samples, respectively. To ensure the robustness and reproducibility of the RNN-LSTM models, each training process was repeated five times using randomly generated datasets based on the above criteria, resulting in 35 models trained by different samples. Additionally, each model was evaluated based on two different metrics: CTC Error and Label Error in order to achieve the best performance of classification. For consistency, identical training parameters were utilized by setting the initial learning (lr) to 0.0001, momentum (β) to 0.9, total epoch number to 80, and using the steepest optimization method. It is important to mention that a key difference between CTC and other temporal classifiers is that CTC does not explicitly segment its input sequences. This has several benefits such as removing the need to locate inherently ambiguous label boundaries (e.g. in speech or handwriting), and allowing label predictions to be grouped together if it proves useful (e.g. if several labels commonly occur together). In any case, determining the segmentation is a waste of modeling effort if only the label sequence is required. One very general way of dealing with structured data would be a hierarchy of temporal classifiers, where the labeling at one level (e.g. letters) becomes inputs for the labeling at the next (e.g. words). Preliminary experiments with hierarchical CTC have been encouraging, and we intend to pursue this direction further. Good generalization is always difficult with maximum likelihood training, but appears to be particularly so for CTC. In the future, we will continue to explore methods to reduce overfitting, such as weight decay, boosting, and margin maximization.

### *3.1. CTC-based Models*

In each training and testing process, various metrics including CTC error, deletions, insertions, label error rate or edit distance which is a character-level error, sequence error rate which is defined by comparing two sequences (words) whether they are identical or not, and substitutions were measured. Figure 8 shows that CTC errors of testing datasets among six fonts had roughly the same trend and error decreased when the combined datasets included all the fonts (this dataset is called “all” in this paper). A comparison using label error rates among all the testing datasets shown in Figure 9 also indicated that character-level error rates were low for all the six fonts. However, “all” datasets had the lowest character-level (edit distance) rates. The other important evaluation metric for OCR engines is the sequence (word-level) error. If a given ground truth and a given predicted sequence are identical, the seqError will be zero; otherwise, it will be assigned by one. Figure 10demonstrates that “all” testing datasets had the best error rates tending to zero, which means almost all word samples were predicted correctly by the trained models. Figures 11, 12 and 13 show the evaluations of the three metrics for training datasets. The same trend as the testing datasets in the results was found. However, the overall error rates were lower as expected.

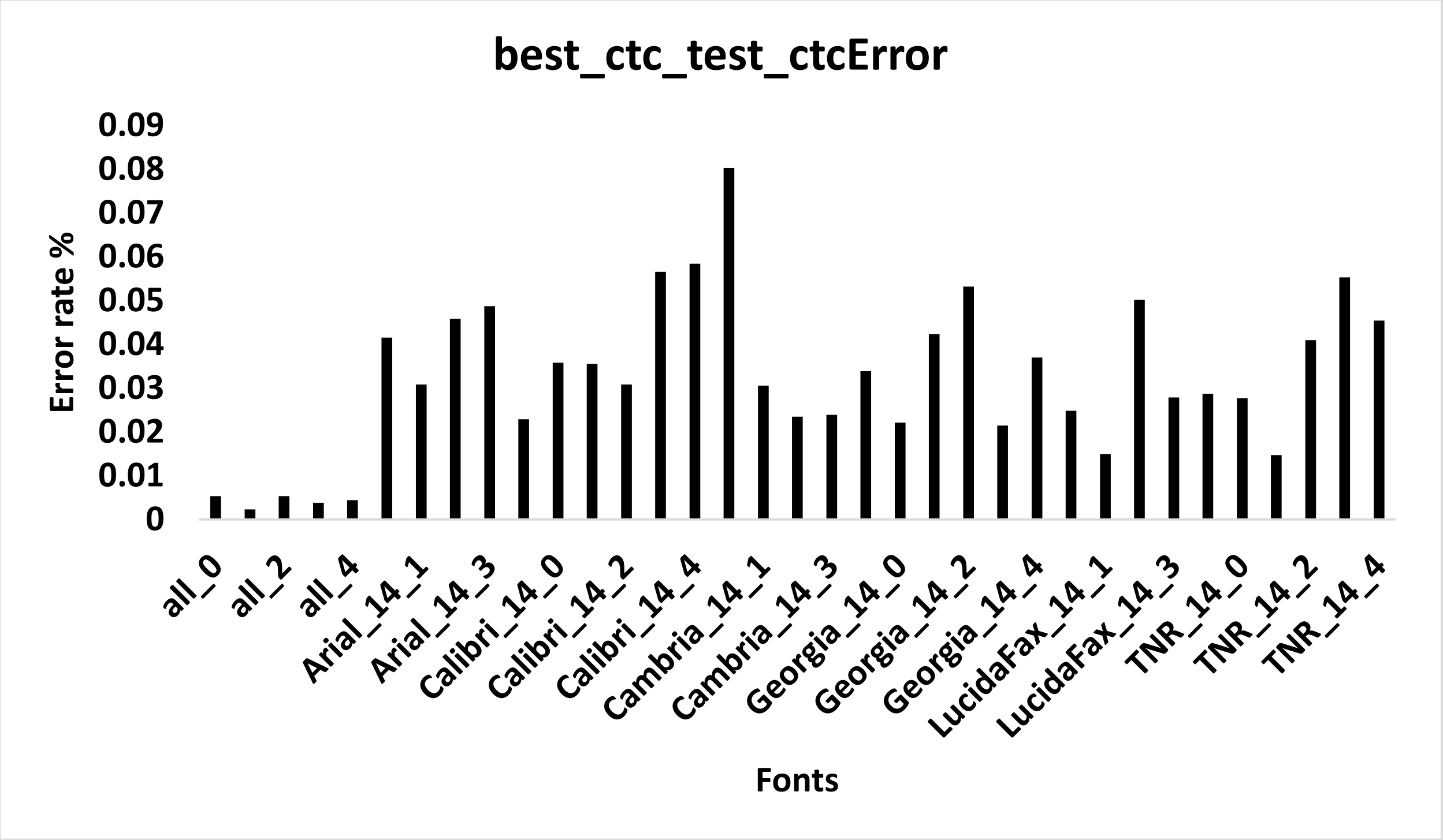

**Figure 8:** Comparison of best models based on CTC criteria and CTC error rates in testing datasets

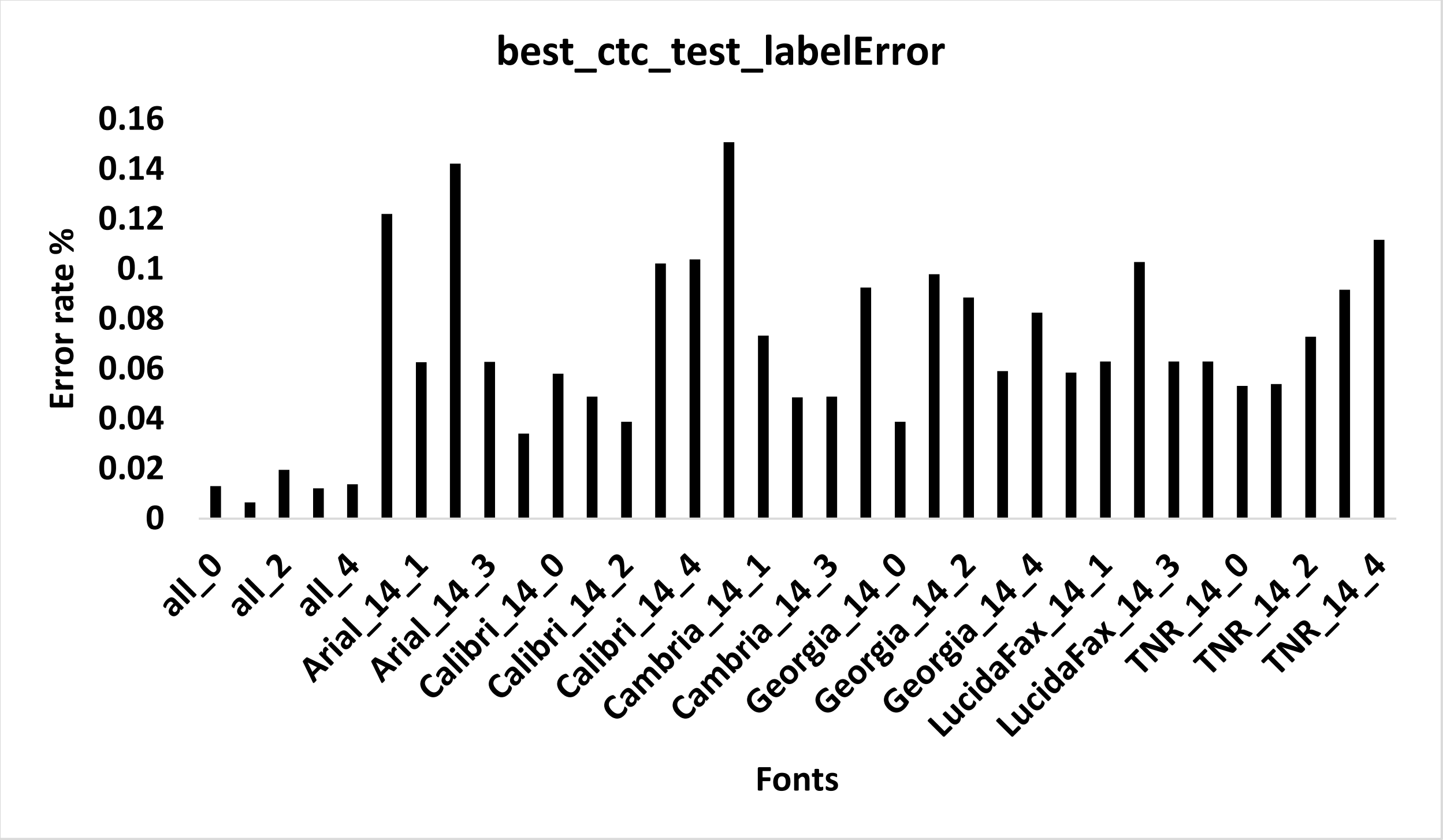

**Figure 9:** Comparison of best models based on CTC criteria and label error rates (edit distance error) in testing datasets

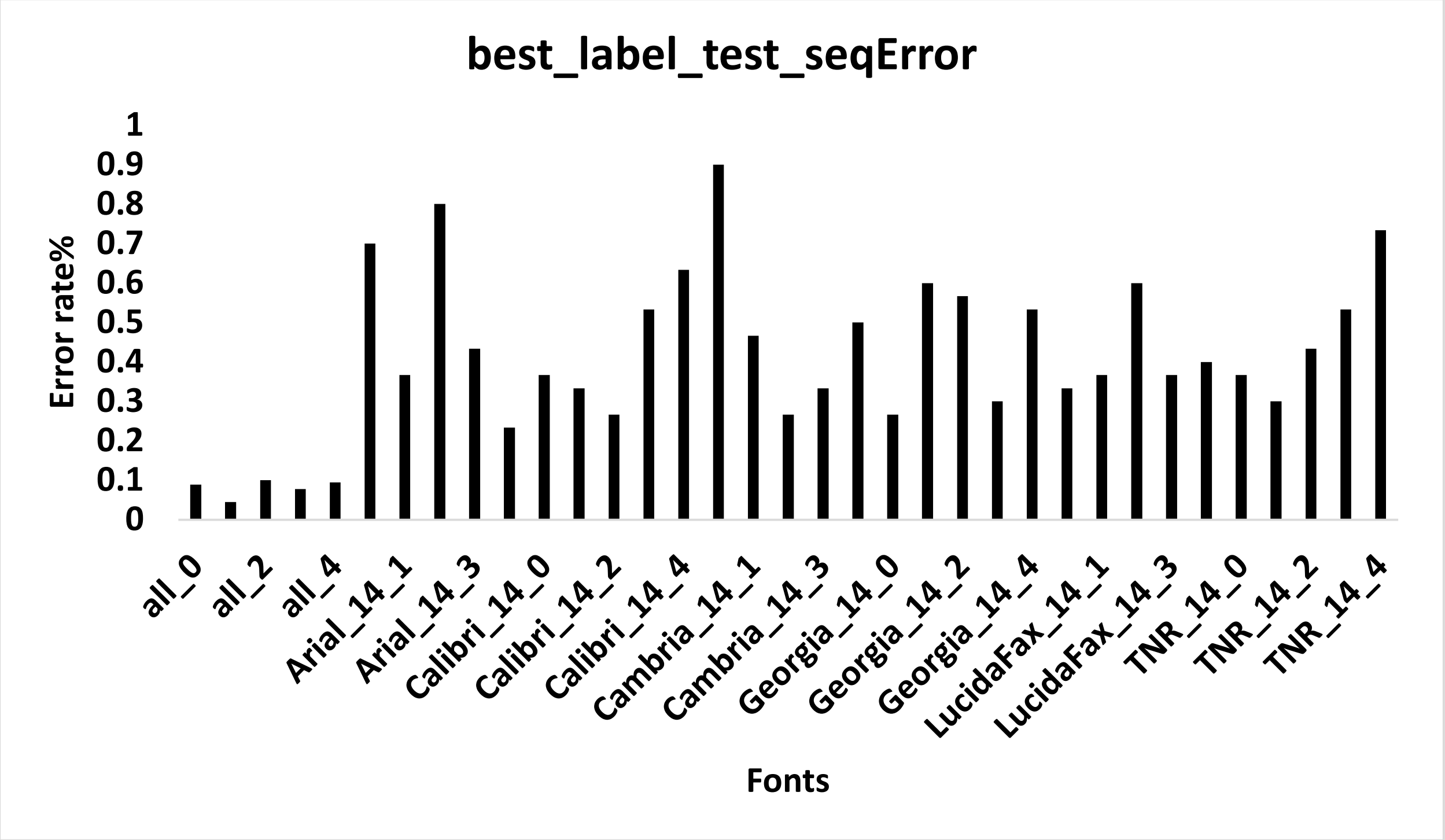

**Figure 10:** Comparison of best models based on CTC criteria and Sequence error rates (word-level error) in testing datasets

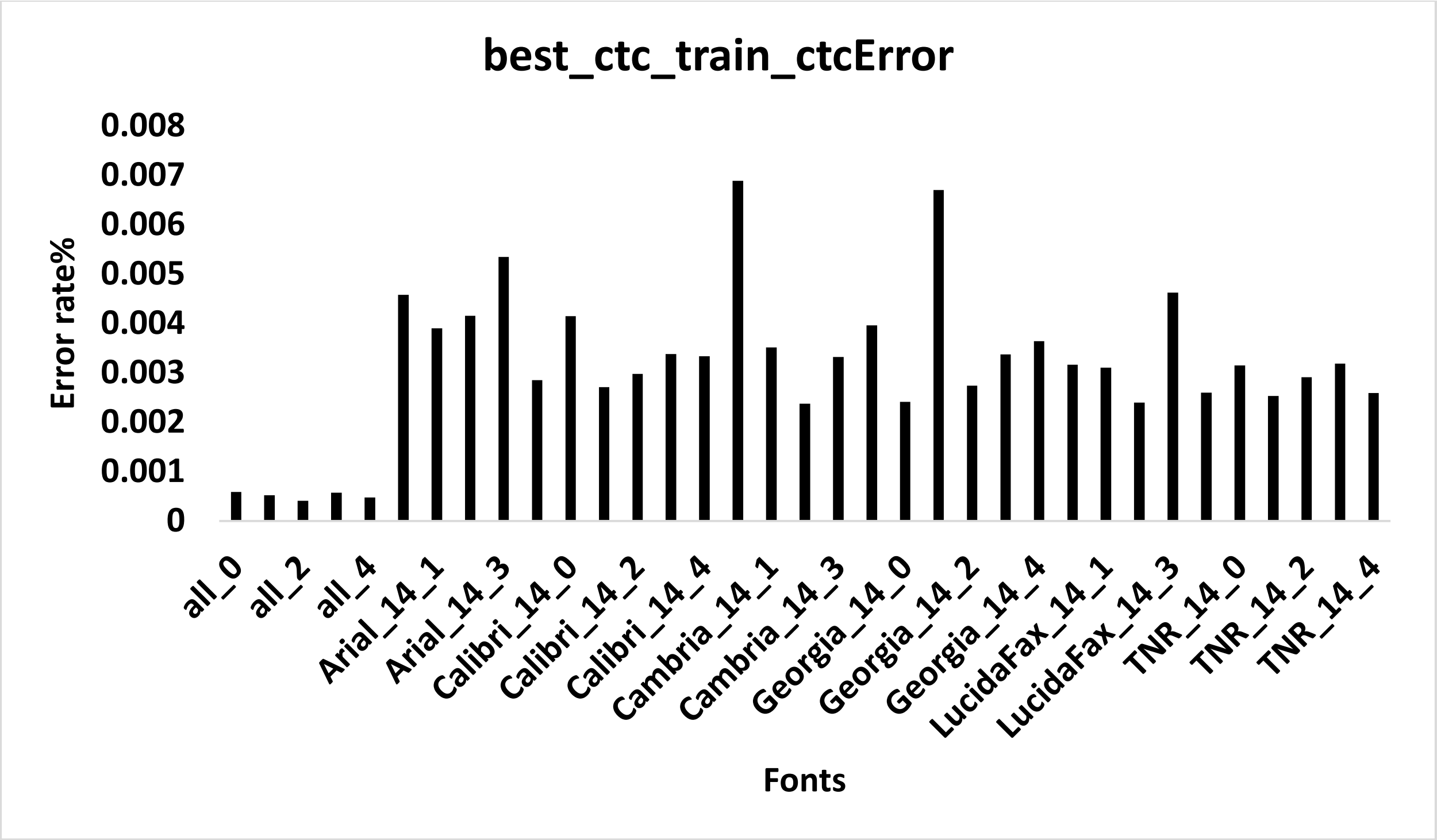

**Figure 11:** Comparison of best models based on CTC criteria and CTC error rates in training datasets

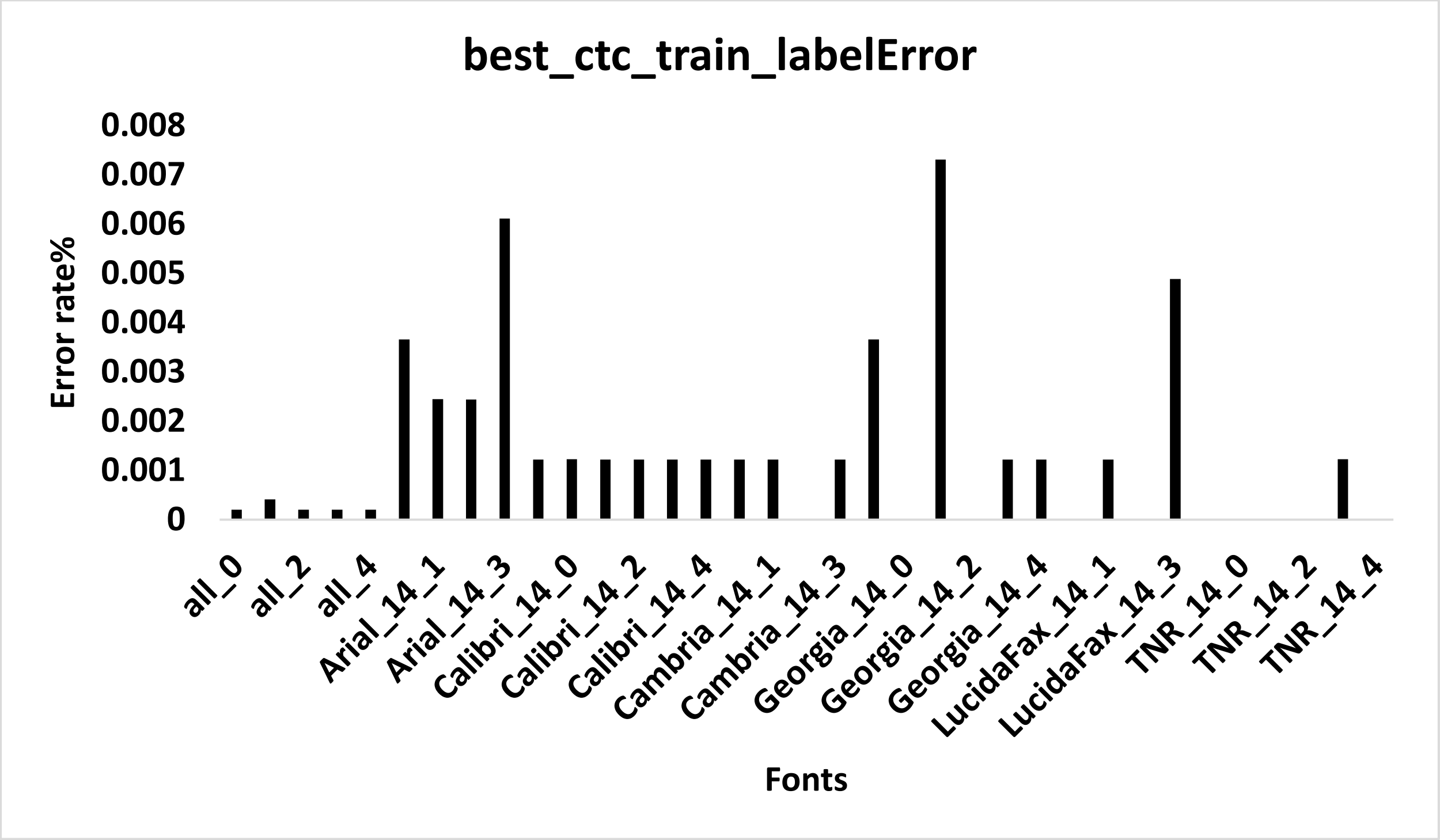

**Figure 12:** Comparison of best models based on CTC criteria and label error rates (edit distance error) in training datasets

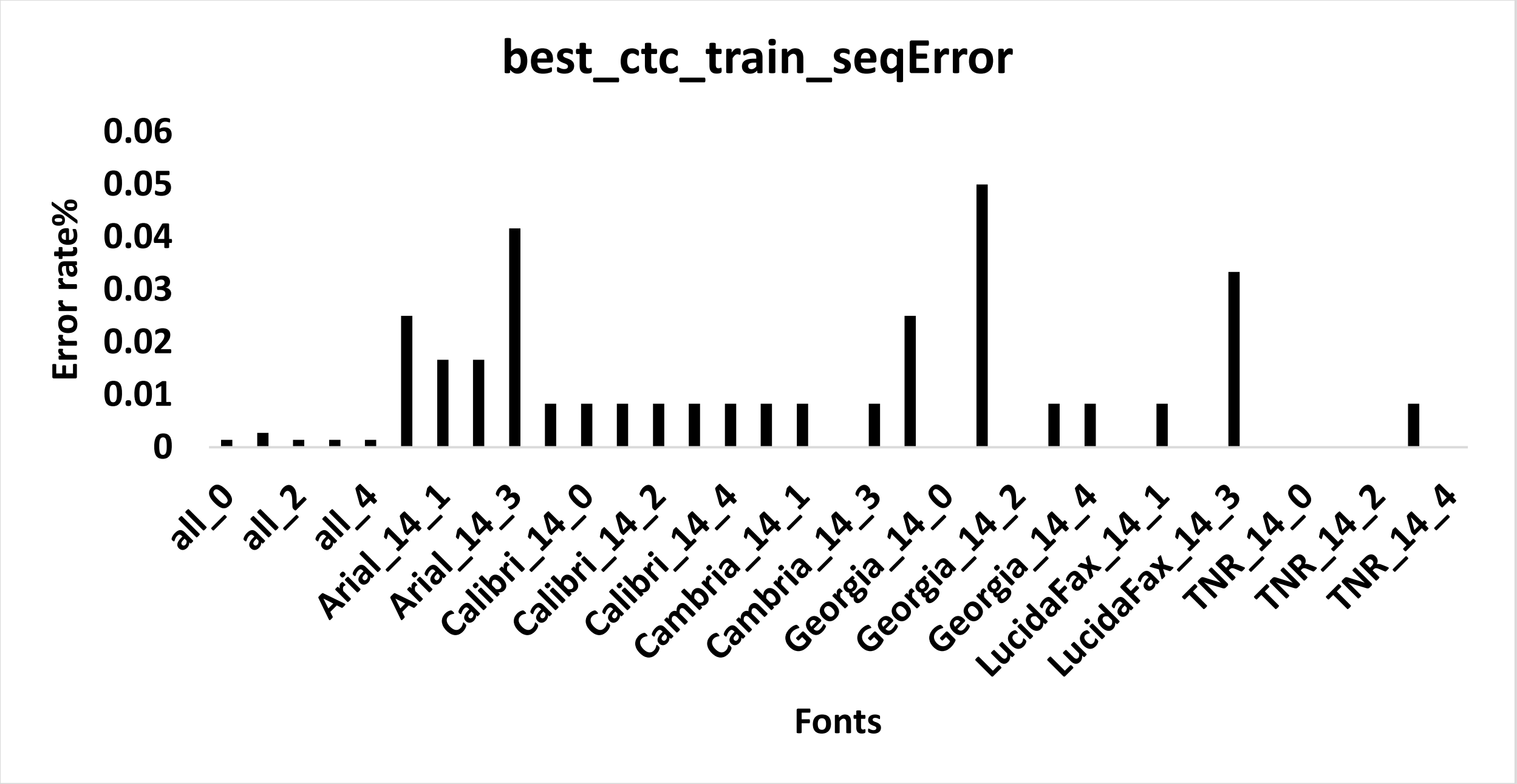

**Figure 13:** Comparison of best models based on CTC criteria and Sequence error rates (word-level error) in training datasets

### *3.2. Label Error-based models*

The best RNN LSTM models based on label errors (edit distance) rates were selected for each training and testing process to evaluate the performance of classification. Figures 14,15 and 16 show that the CTC, label error, and sequence error rates for all of the fonts' testing samples are extremely low. Also, the error rates for "all" testing dataset tends to zero which means the predication of the French words was almost perfectly carried on. The similar comparison shown in Figures 17, 18, 19 among the metrics mentioned above illustrates that the training process perfectly performed.

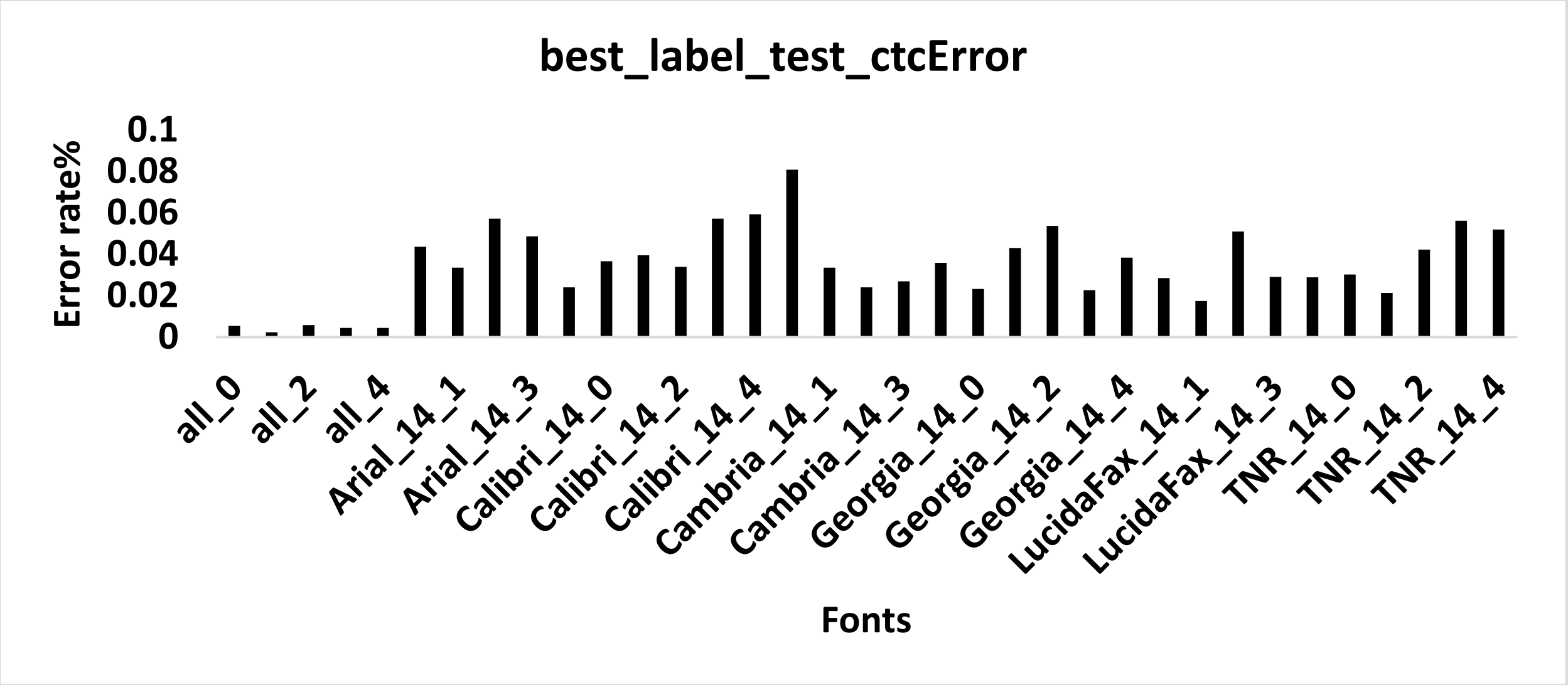

**Figure 14:** Comparison of best models based on label error criteria and CTC error rates in testing datasets

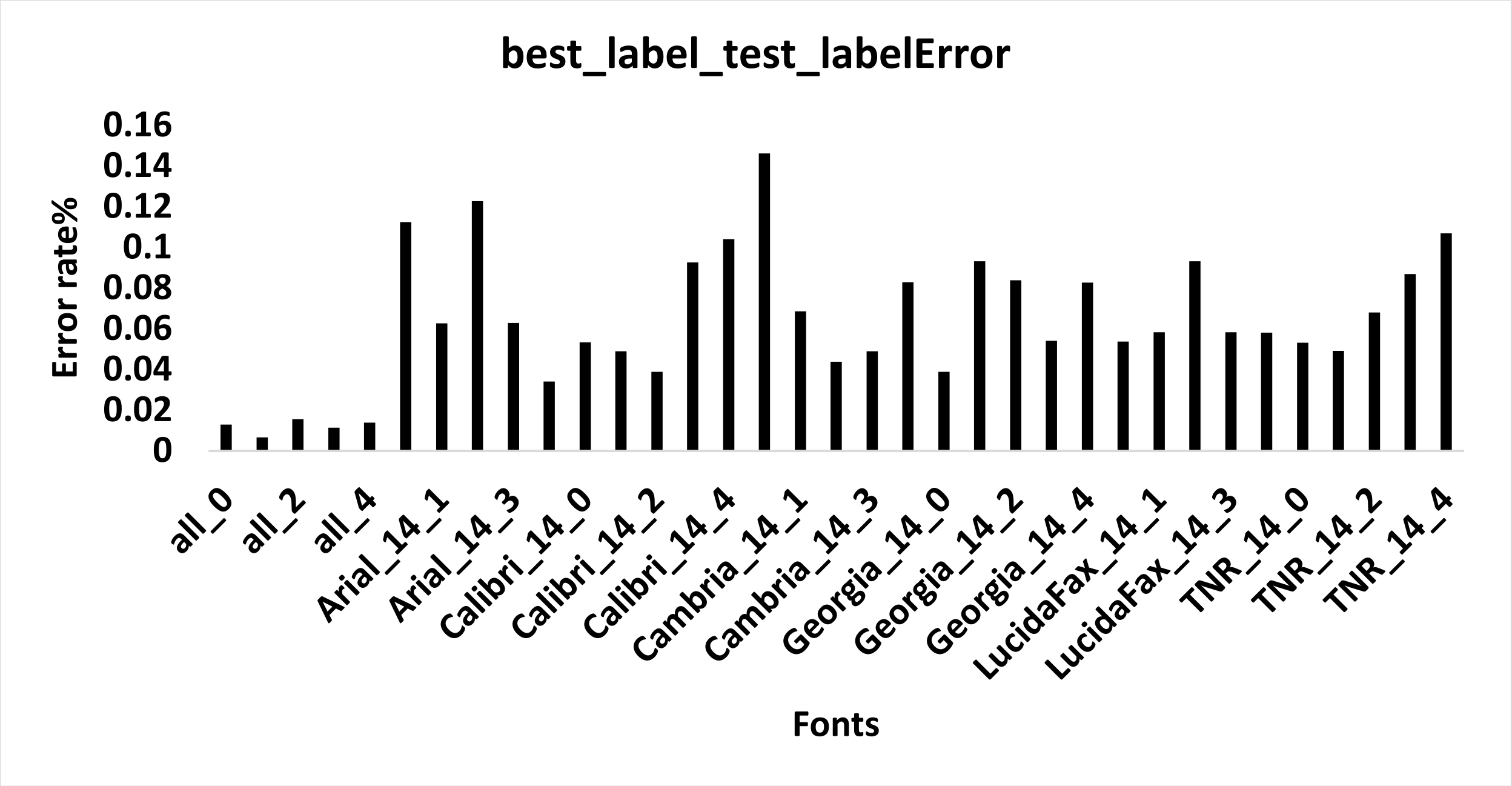

**Figure 1:** Comparison of best models based on label error criteria and label error rates (edit distance error) in testing datasets

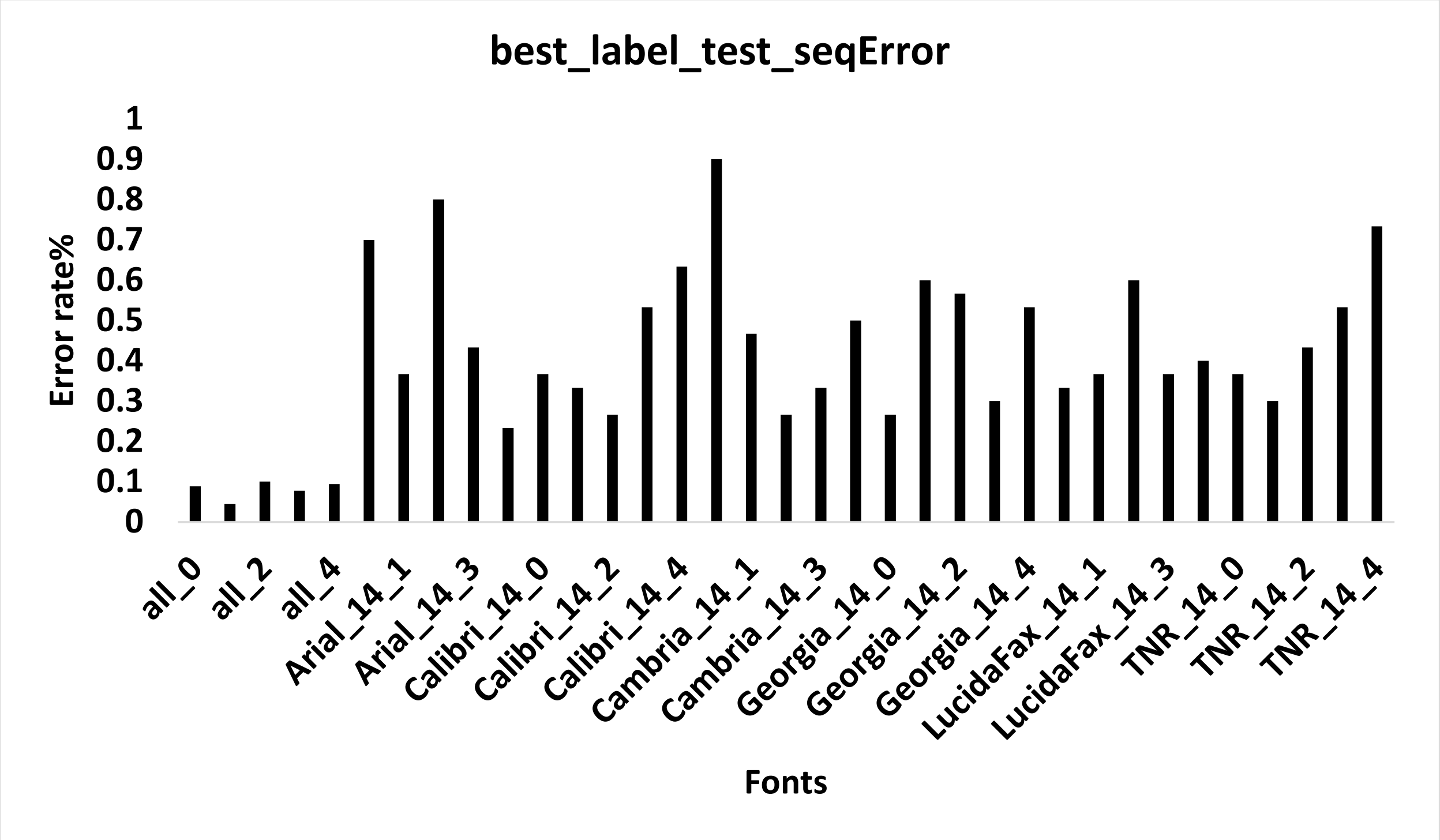

**Figure 2:** Comparison of best models based on label error criteria and Sequence error rates (word-level error) in testing datasets

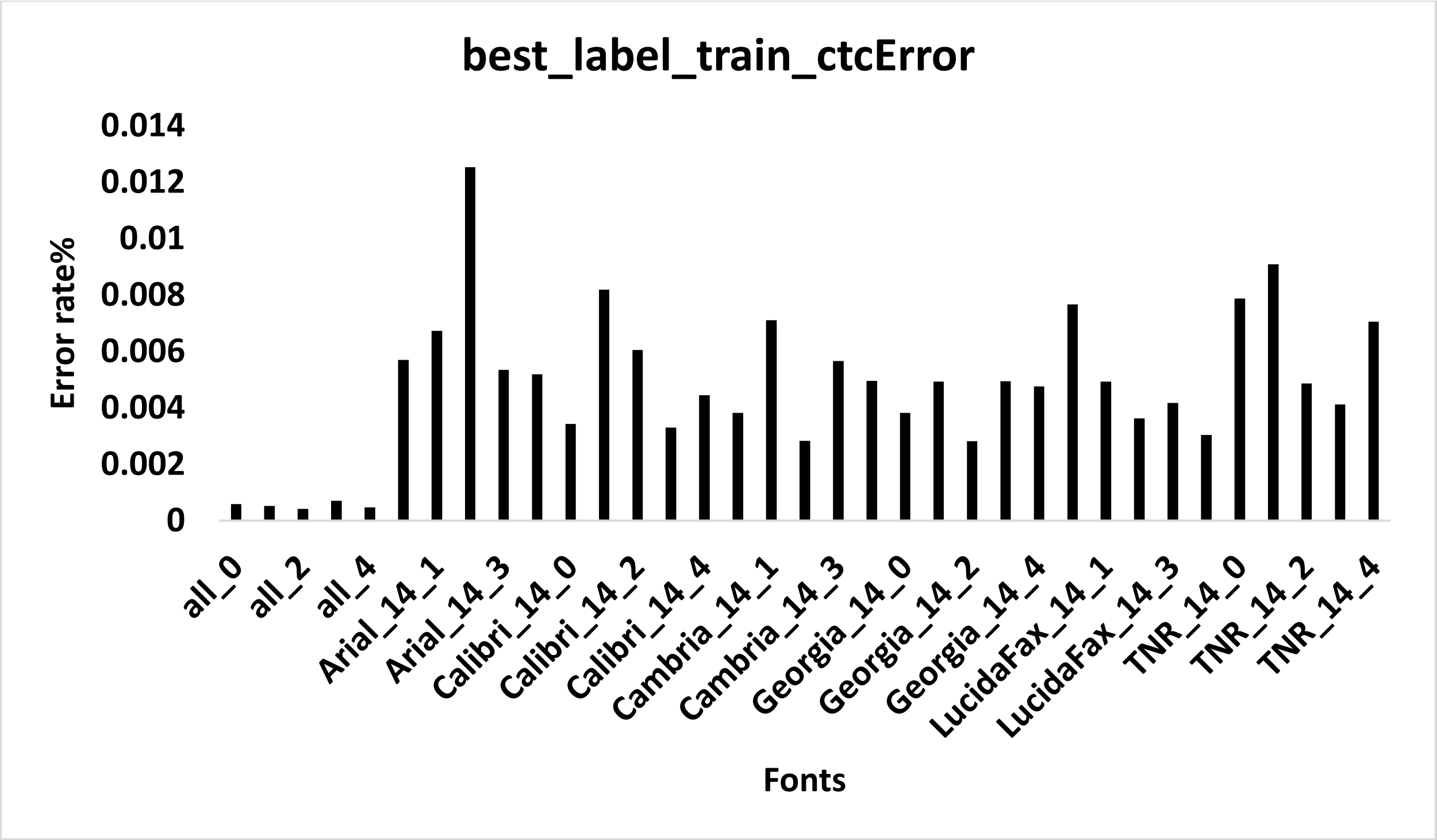

**Figure 3:** Comparison of best models based on label error criteria and CTC error rates in training datasets

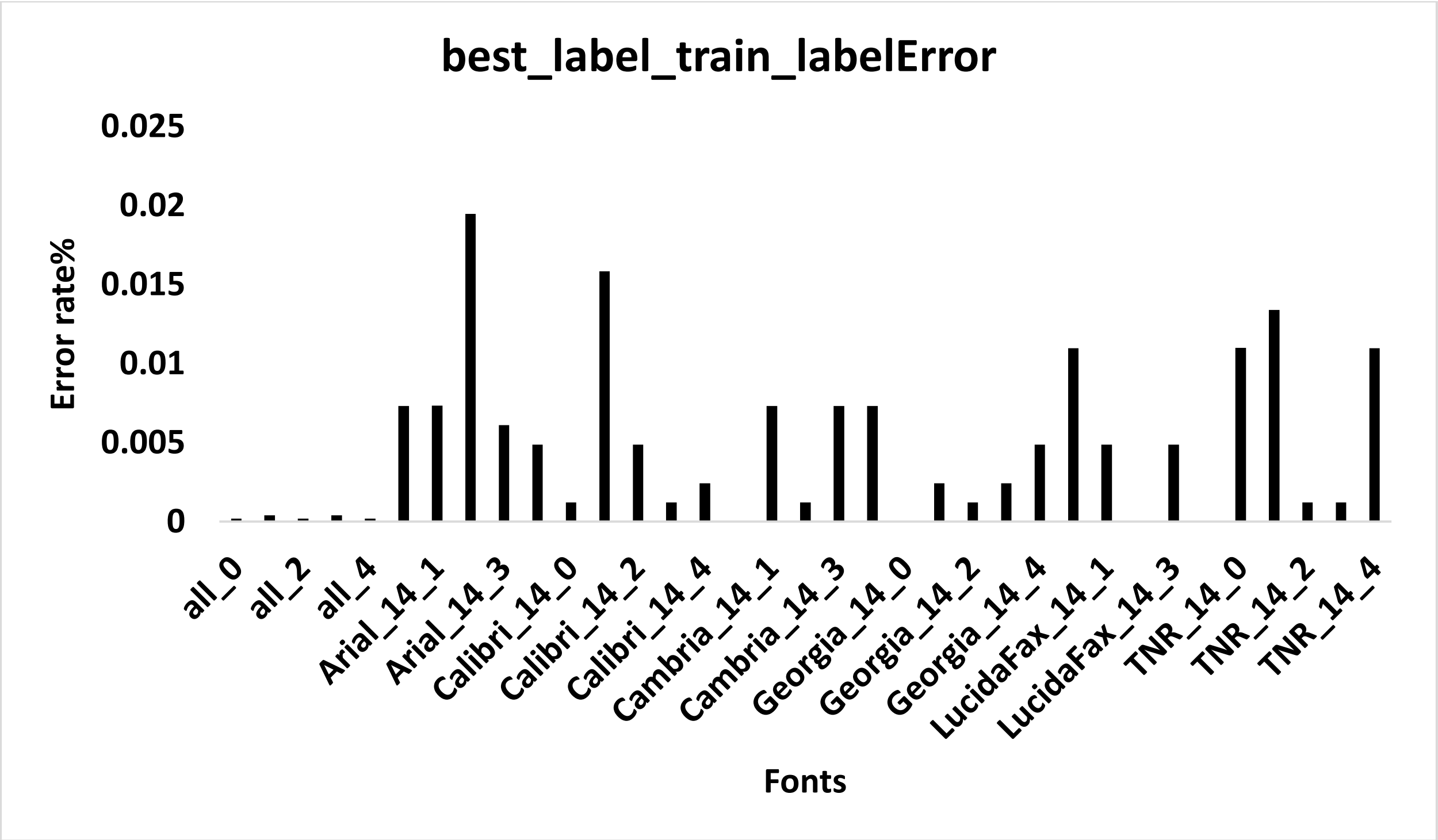

**Figure 4:** Comparison of best models based on label error criteria and label error rates (edit distance error) in training datasets

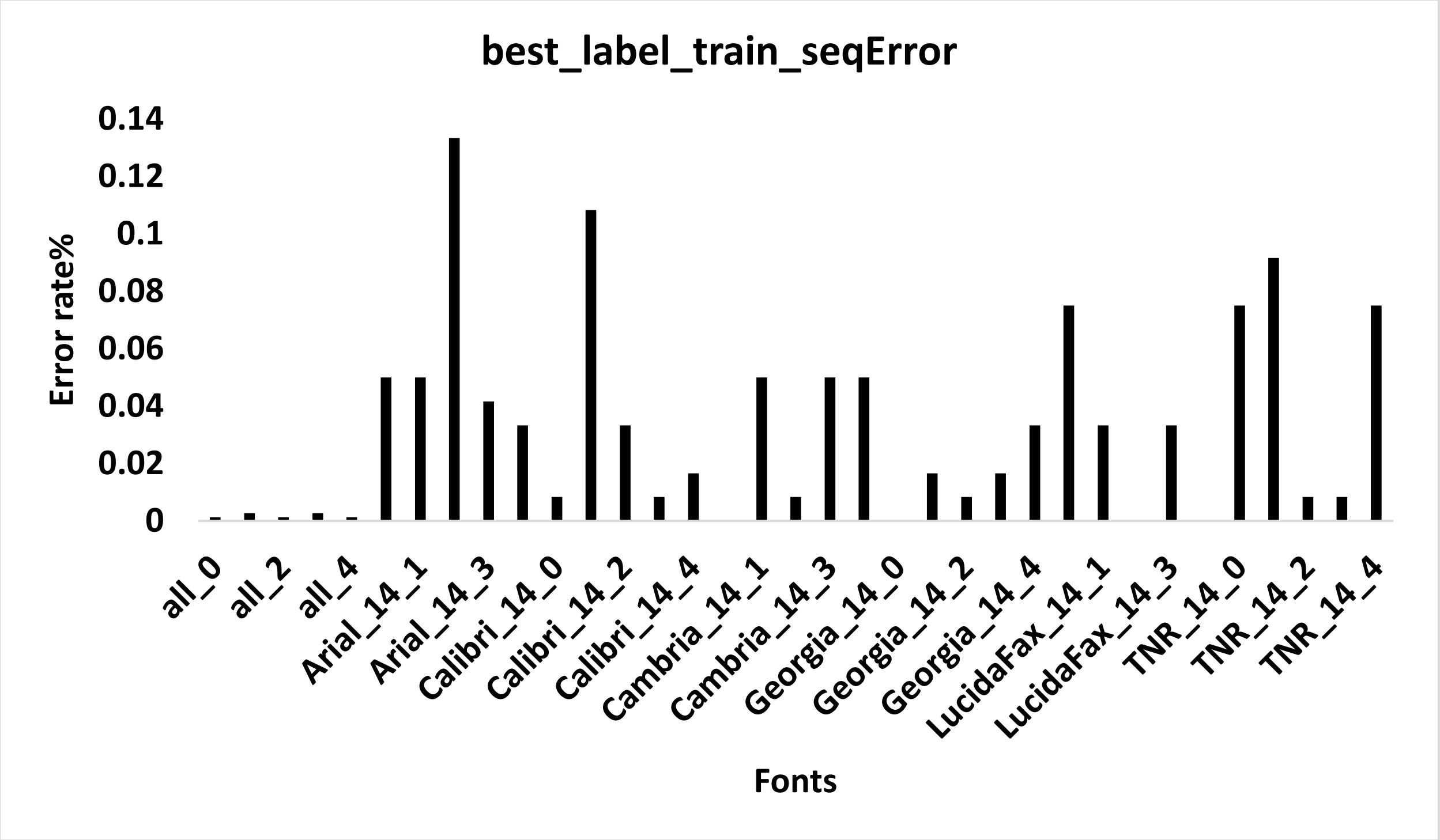

**Figure 5:** Comparison of best models based on label error criteria and Sequence error rates (word-level error) in training datasets

### *3.3. Best Models Comparison*

The training process of each font and "all" datasets was repeated five times using different randomly selected samples as mentioned earlier. In order to find the best model within a group, the models with the lowest error rates according to CTC and label error rates were selected. Table 1 indicates that the combined dataset "all" had the lowest error rates in both evaluation methods.

The label error of the best model reached 0.0065% and the sequence error rate was 0.044%. In other words, the trained RNN-LSTM model correctly predicted over 99.56% of the French words used for testing. Figures 20 and 21 show the performance of classification for all fonts based on two approaches and indicates that the results from both methods are highly correlated.

As mentioned above, the "all" training and testing datasets included all samples of six fonts. Although the variation among fonts might affect the performance of classification, an increase of six fold in the number of samples decreased dramatically from 0.366% (worst-case scenario) to 0.044% for both label error and sequence error in both CTC and label error-based evaluation, which images the effect of increasing the number samples on deep learning training process. In addition, the average of five-times repetitions of each training process was shown in Table 2.

The results indicate that in both character and sequence level, the performance of classification is almost perfect and all French words in testing process were predicted correctly.

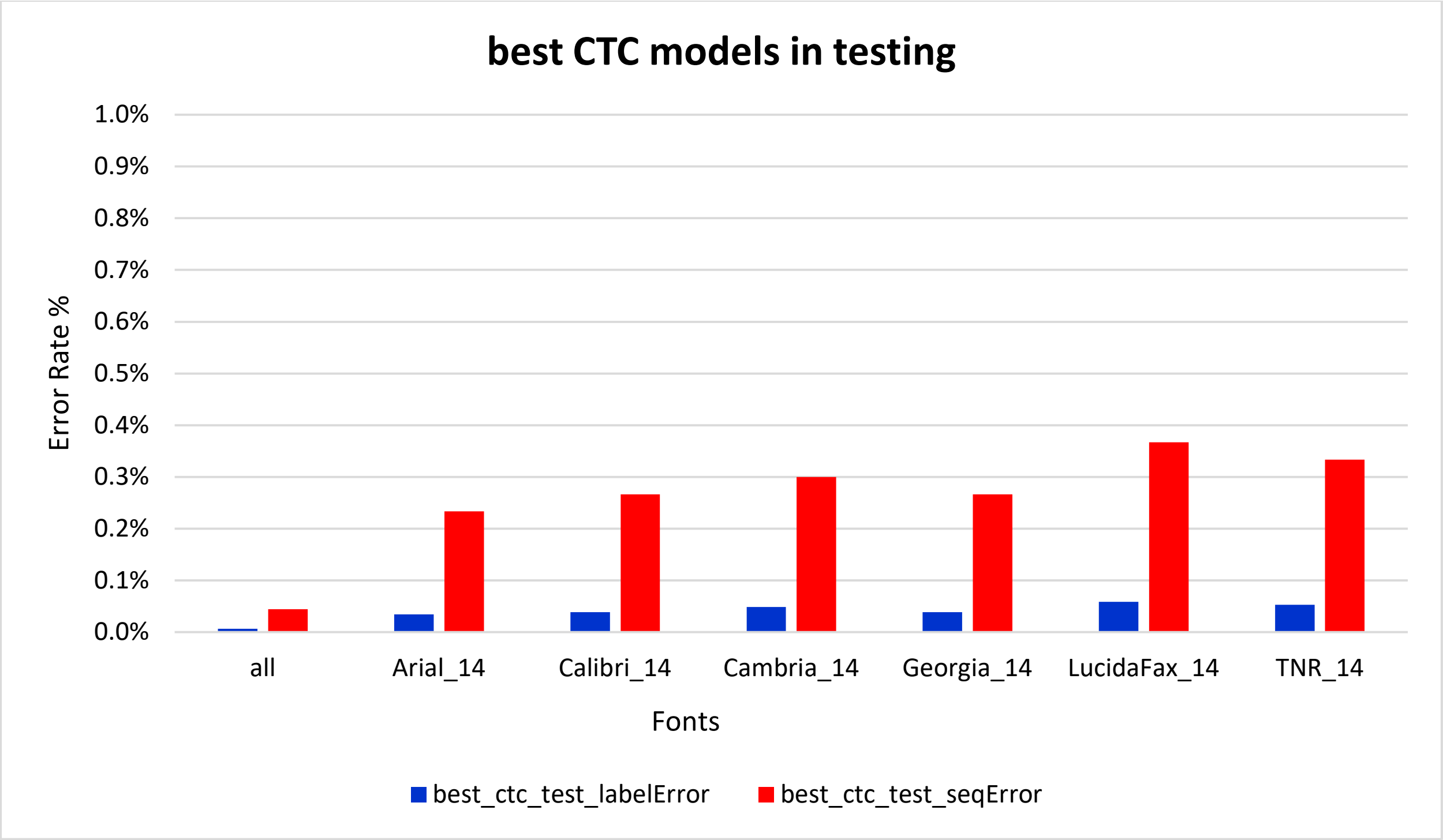

**Figure 6:** Comparison of best models using CTC criteria among five-times repetitions for each testing datasets

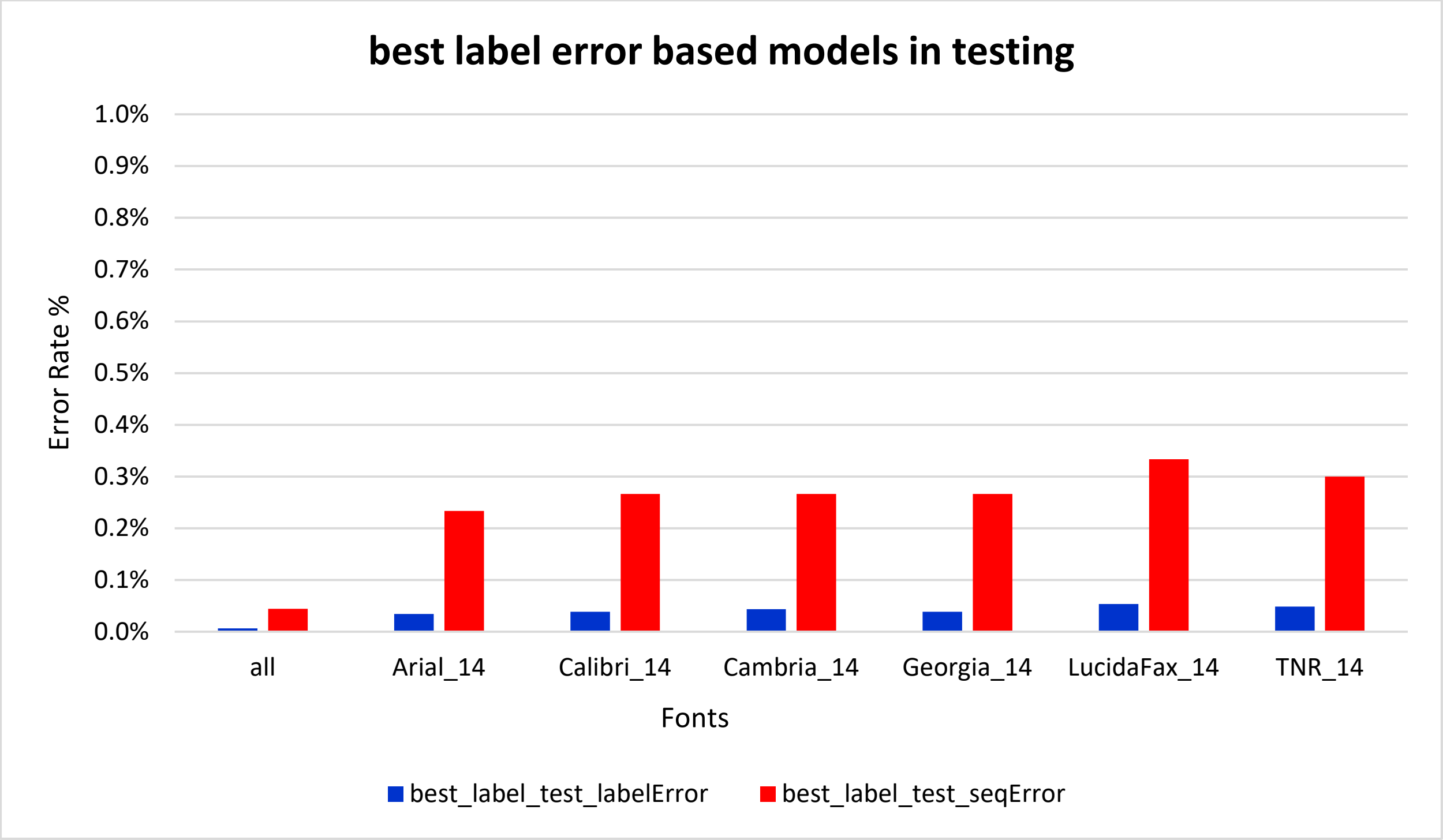

**Figure 7:** Comparison of best models using label error criteria among five-times repetitions for each testing datasets

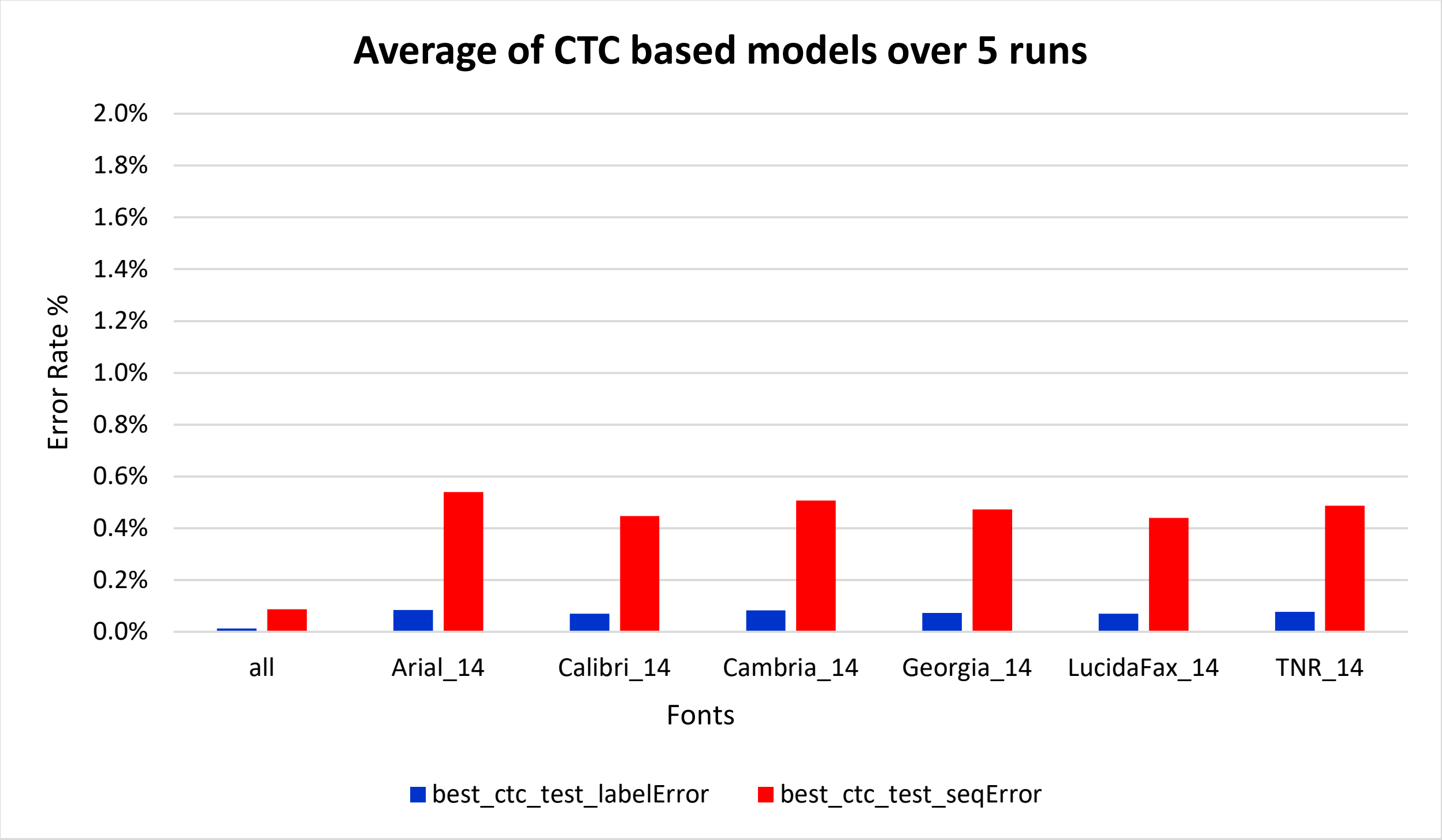

**Figure 8:** Comparison of averaged models for five-times repetitions using CTC criteria for each testing datasets

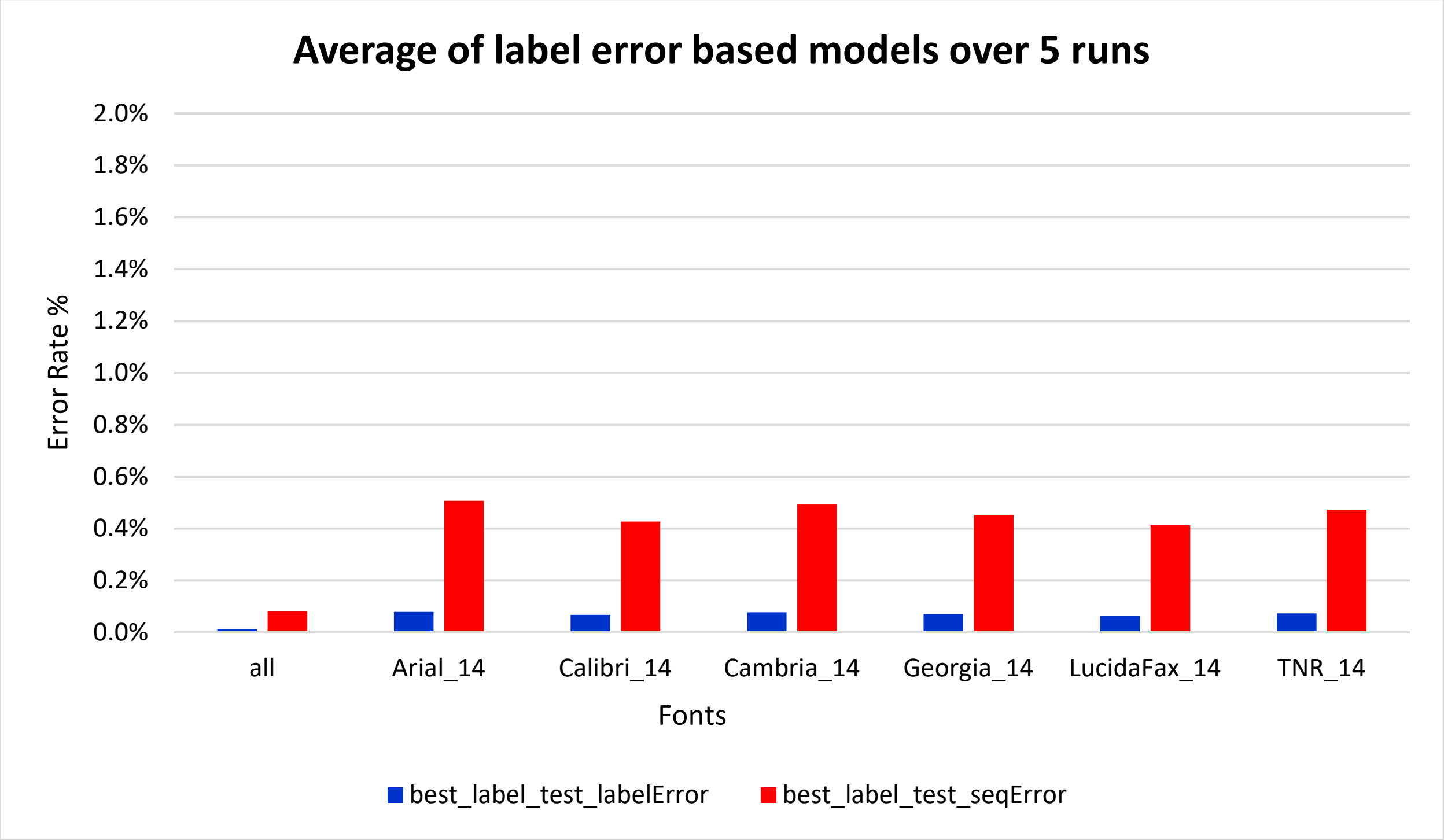

**Figure 9:** Comparison of averaged models for five-times repetitions using label error criteria for each testing datasets

**Table 1:** The best models based on CTC and label error criteria are shown in this table. “All” dataset including six fonts showed the best classification performance for recognizing French OCR samples. As explained, providing sufficient samples to train RNN-LSTM model is necessary. Also, data variation, e.g. adding more fonts and samples, is increased in the training process.

| Test datasets | best_ctc_ labelError | best_ctc_seqError | best_label_ labelError | best_label_ seqError |
|---|---|---|---|---|
| all | 0.00650% | 0.04444% | 0.00650% | 0.04444% |
| Arial_14 | 0.03411% | 0.23333% | 0.03411% | 0.23333% |
| Calibri_14 | 0.03876% | 0.26667% | 0.03876% | 0.26667% |
| Cambria_14 | 0.04866% | 0.30000% | 0.04380% | 0.26667% |
| Georgia_14 | 0.03887% | 0.26667% | 0.03887% | 0.26667% |
| LucidaFax_14 | 0.05851% | 0.36667% | 0.05363% | 0.33333% |
| TNR_14 | 0.05314% | 0.33333% | 0.04900% | 0.30000% |

**Table 1:** The averaged models of five-time repetitions based CTC and label error criteria are described. The results show an identical trend as in Table 1.

| Test datasets | best_ctc_labelError | best_ctc_seqError | best_label_labelError | best_label_seqError |
|---|---|---|---|---|
| all | 0.01299% | 0.08778% | 0.01202% | 0.08111% |
| Arial_14 | 0.08480% | 0.54000% | 0.07892% | 0.50667% |
| Calibri_14 | 0.07039% | 0.44667% | 0.06747% | 0.42667% |
| Cambria_14 | 0.08290% | 0.50667% | 0.07803% | 0.49333% |
| Georgia_14 | 0.07342% | 0.47333% | 0.07047% | 0.45333% |
| LucidaFax_14 | 0.07006% | 0.44000% | 0.06422% | 0.41333% |
| TNR_14 | 0.07666% | 0.48667% | 0.07278% | 0.47333% |

### *3.4. Early Stopping versus Fixed Epochs*

Training any classifiers such as RNN-LSTM requires a policy or criteria to stop at certain stage. Several methodologies already exist in machine learning to define how and where any training process should be terminated. As various groups in the field of computer vision and machine learning have already shared their investigation and results with researchers, an optimal value is considered the number of epochs for training a given classifier. This approach enables researchers to meaningfully compare various trained models with respect to different error rates, such as edit distance or sequence error in this study. Figure 24 shows a comparison between RNN-LSTM trained models against “all” training testing datasets. The number of epochs was set to 80 to compare all the models equally. The results demonstrate that the error rates tend to zero after around 40 epochs. It would suggest that training a RNN-LSTM model using printed character samples usually converges after 40 epochs and further training might not improve validation results. Another method for terminating training process is using early stopping criteria. RNNLIB provides criteria in which the training process continues while the algorithm keeps tracking the last 20 epochs. No improvement in edit distance error rate in the past 20 epochs cause the training process to be stopped. This methodology may cause a very long training process but may result a better-trained model.

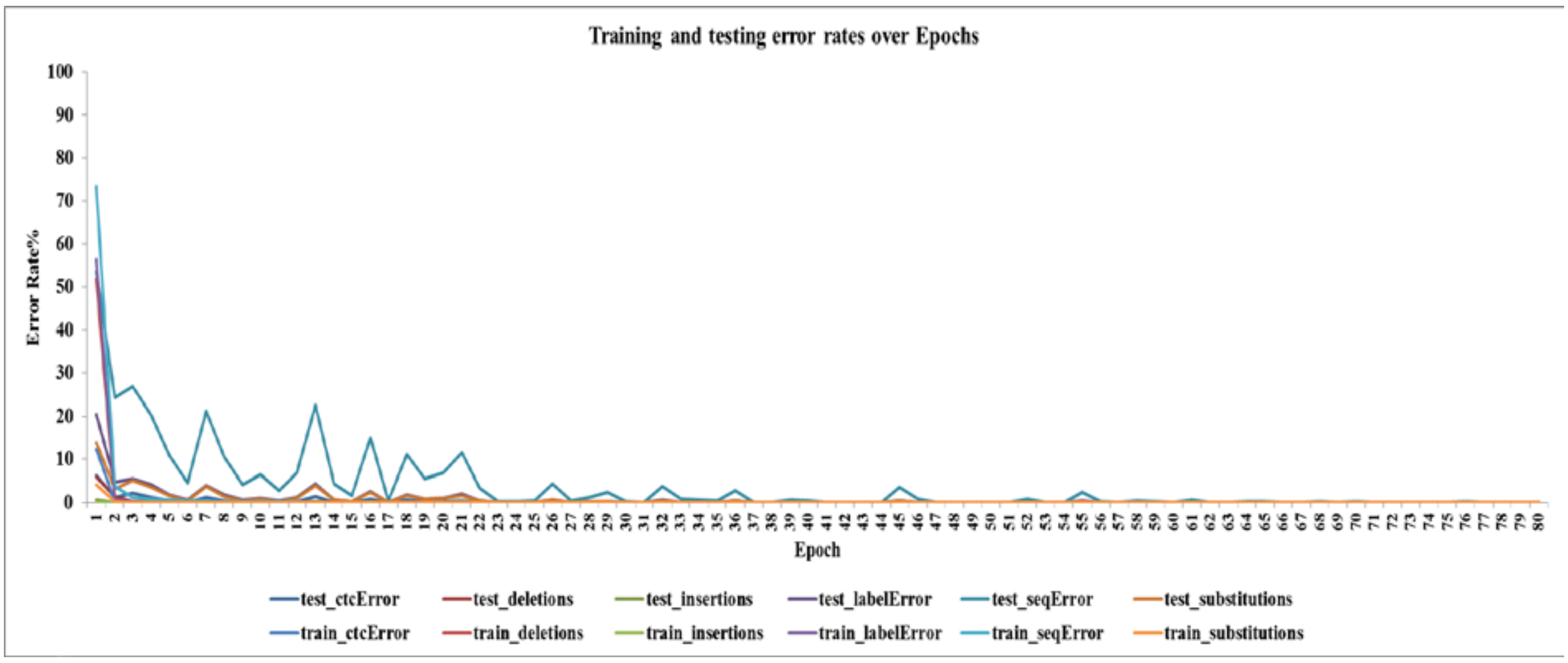

**Figure 24:** Comparison of various error rates between RNN-LSTM trained models against "all" training and testing datasets

## 4. Conclusion

Optical character recognition is a fundamental concept in computer vision that is considered established technology. Recent research has shown that using convolutional neural networks significantly improved the accuracy rates of classification. However, a state-of-art design of recurrent neural networks through long-short term memory blocks (RNN-LSTM) presented a more advanced, meaningful, and accurate architecture for solving the classical question (OCR). In this experiment, seven datasets including samples generated for six fonts among one dataset of all samples using the top 5,000 French words in public data were utilized to train and test an RNN-LSTM by RNNLIB. The accuracy rates of these seven datasets were extremely high and the "all" dataset including all the samples showed the highest accuracy rate of around full accuracy. This study showed a successful design of OCR pipeline using RNN-LSTM in which not too many samples were utilized for training, but no explicit feature extraction module was used. However, the study demonstrated that providing sufficient numbers of samples enable the classifier to be trained more accurately.

## 5. Disclosure Statement

The authors declare that there is no conflict of interest regarding the publication of this manuscript. This work was completed only for research purpose based on author's funding and interest and has no conflict with any companies or organizations research or products.